\documentclass{article}
\usepackage[square,numbers]{natbib}
\usepackage{authblk}
\usepackage{charter}
\usepackage{fullpage}
\usepackage{bbm}

\usepackage{hyperref}       % hyperlinks
\usepackage{url}            % simple URL typesetting
\usepackage{booktabs}       % professional-quality tables
\usepackage{amsfonts}       % blackboard math symbols
\usepackage{nicefrac}       % compact symbols for 1/2, etc.
\usepackage{microtype}      % microtypography
\usepackage[table]{xcolor}         % colors
\usepackage{amsmath,amssymb,amsthm,amsfonts}
\usepackage{latexsym,bbm,graphicx,float,mathtools}
\usepackage{algorithmic}
\usepackage[export]{adjustbox}
\usepackage{capt-of}% or \usepackage{caption}
\usepackage{booktabs}
\usepackage{parskip}

\usepackage[utf8]{inputenc}

\usepackage{fvextra}

\DefineVerbatimEnvironment{AutoVerb}{Verbatim}{
  fontsize=\small,           % \small, \footnotesize, \scriptsize 等
  breaklines=true,           % 在空格处断行
  breakanywhere=true,        % 在任意字符处断行
  breaksymbol=\tiny\ensuremath{\hookrightarrow}, % 断行提示符
  frame=single,              % 单线边框（可选）
  numbers=left,               % 左侧行号（可选）
  commandchars=\\\{\}
}

\usepackage[noabbrev]{cleveref}
\usepackage{caption}
\usepackage{subcaption}
\usepackage{transparent}
\usepackage[inline]{enumitem}
\usepackage{multirow}
\usepackage{multicol}
\usepackage{algorithm}
\usepackage{xspace}

\usepackage{abstract}

\usepackage[subtle, mathdisplays=tight, charwidths=normal, leading=normal]{savetrees}

\title{\textbf{OpenTable-R1: A Reinforcement Learning Augmented Tool Agent for Open-Domain Table Question Answering}}

\author{
    Zipeng Qiu\\
}

\date{}

\begin{document}
\maketitle

\begin{abstract}
\vspace{0.75em}
Open-domain table question answering traditionally relies on a two-stage pipeline: static table retrieval followed by a closed-domain answer. In contrast, we propose an end-to-end agentic framework that embeds multi-turn tool calls—using a BM25+-based search API and a SQLite SQL executor—directly into a large language model. To further adapt a compact 4B-parameter model, we introduce a two-stage fine-tuning process: supervised cold-start on easy questions, then Async GRPO reinforcement learning on harder cases with LoRA adapters and a rollout buffer. This unified approach enables the model to jointly retrieve, reason, and execute queries, yielding a dramatic accuracy improvement from single-digit zero-shot performance to over 86\% exact match on a held-out test set. Our results underscore the effectiveness of integrating structured tool calls with targeted RL fine-tuning for scalable, accurate table QA. The code is available at \url{https://github.com/TabibitoQZP/OpenTableR1}.

\end{abstract}

%\begingroup
%\renewcommand\thefootnote{}

%\addtocounter{footnote}{-1}
%\endgroup

\section{Introduction}

Large-scale language models have demonstrated remarkable proficiency across a wide array of downstream tasks, from summarization to code generation. However, open-domain table question answering presents a unique challenge: an immense pool of heterogeneous tables cannot be serialized directly into the model’s context. Traditional pipelines therefore decouple retrieval and reading, employing static table-retrieval algorithms to identify candidate tables before invoking a closed-domain QA model on the selected subset. While effective to an extent, this approach inherently lacks flexibility and feedback between retrieval and reasoning.

Recent advances in LLM ``agentic" behavior—wherein the model interacts with external tools via API calls—offer a promising alternative \cite{grattafiori2024llama,qwen2025qwen25technicalreport,yang2025qwen3}. By exposing dedicated table-search and SQL-query interfaces, we can leverage the model’s inherent reasoning capabilities to perform end-to-end open-domain table QA. Moreover, reinforcement learning (RL) methodologies have been shown to significantly boost model performance on tasks requiring sequential decision-making \cite{schulman2017proximal,rafailov2023direct,shao2024deepseekmath}. Combining tool-enabled interaction with RL therefore offers a powerful framework for addressing open-domain table QA in a more integrated and adaptive manner.

Despite this promise, our experiments reveal that naïvely equipping LLMs with table-search and SQL-execution tools yields only marginal gains over static pipelines. The model often fails to learn when and how to invoke each tool effectively, leading to suboptimal retrieval and query strategies. This observation highlights the critical need for targeted enhancement of the model’s tool-usage capabilities, a topic that has received scant attention in the context of open-domain table QA.

To date, most work on open-domain table QA has focused exclusively on improving retrieval quality, under the assumption that better retrieval directly translates to higher end-task accuracy. These static retrieval-then-QA systems offer no mechanism for recovery when the initial table selection is incorrect, and multi-agent architectures—while capable of coordinating keyword search and structured querying—incur substantial token and computational overhead. By contrast, modern language models now possess far more robust inference and tool-calling abilities, creating an opportunity to revisit open-domain table QA with an end-to-end, tool-centric paradigm.

In this paper, we make the following contributions:

\begin{itemize}
    \item \textbf{Comprehensive evaluation of LLM tool usage:} We conduct extensive tool‐invocation tests across multiple model sizes, comparing how different architectures leverage table‐search and SQL‐query tools.
    \item \textbf{Reinforcement-learning-based tool‐invocation policy:} We propose an RL framework that trains the LLM to balance tool-calling cost against QA reward, optimizing end-to-end accuracy.
    \item \textbf{State‐of‐the‐art results on open WikiTable QA:} Using our enhanced approach with a simple retrieval algorithm, we achieve over 86\% exact match on the Open WikiTable benchmark—substantially higher than both static retrieval baselines and LLMs with native, unsupervised tool calls.
    % \item \textbf{Ablation across retrieval strategies:} We analyze how different table-retrieval methods interact with our RL-augmented tool policy, elucidating the trade-offs between retrieval quality, tool-calling frequency, and overall accuracy.
\end{itemize}

\section{Methodology}

In this section, we will introduce our tool call implementation and the adaptation of GRPO algorithm.

\subsection{Tool Call Implementation}

To enable our open-domain table QA agent to interact effectively with large language models (LLMs), we design and expose two core tool APIs: a \texttt{search} tool for table retrieval and a \texttt{code\_interpreter} tool for executing SQL queries.

\paragraph{Table Retrieval Tool}  
The \texttt{search} tool accepts two parameters, \texttt{keywords} (a free-form query string) and \texttt{top\_k} (an integer), and returns the top~$k$ tables most relevant to the query. Internally, we index all tables in the corpus and apply a BM25+ matching algorithm for initial experiments, owing to its simplicity and efficiency. In future work, we will compare BM25+ against advanced dense retrieval and neural ranking models to compare the performance.

\paragraph{SQL Query Tool}  
The \texttt{code\_interpreter} tool provides a sandboxed interface for executing arbitrary SQL statements over our tabular dataset. All tables are preloaded into a SQLite database, enabling lightweight and dependency-free query execution. The tool takes a single argument, \texttt{sql\_query}, which should be a valid SQL statement. Upon invocation, the tool executes the statement and returns the result set in a specific format.

\begin{table}[!htbp]
  \centering
  \begin{tabular}{@{}llll@{}}
    \toprule
    \textbf{Tool Name} & \textbf{Argument}    & \textbf{Type} & \textbf{Description} \\
    \midrule
    \texttt{search}            & \texttt{keywords} & str           & Query string for table retrieval. \\
                               & \texttt{top\_k}   & int           & Number of tables to return. \\
    \midrule
    \texttt{code\_interpreter} & \texttt{sql\_query} & str         & Single-line SQL statement to execute. \\
    \bottomrule
  \end{tabular}
  \caption{APIs exposed to the LLM for table retrieval and SQL execution.}
  \label{tab:tool_apis}
\end{table}

\paragraph{Prompt Design for Tool Invocation}  
% Following the mainstream paradigm of multi-turn dialogue for tool-enabled QA, we first instruct the model in the system message that it is provided with specific tools and how to use them.  In each planning turn, the LLM decides which tool to call and with what arguments; in each execution turn, it emits only the tagged invocation.  To support this, we include concise descriptions of the \texttt{search} and \texttt{code\_interpreter} tools—listing their names, expected arguments, and output schema—directly in the prompt.  We then require that every tool call be formatted as a single-line, JSON-like function call wrapped in lightweight XML-style tags around the tool name.  This explicit tagging ensures the model clearly distinguishes between natural-language reasoning and executable code, and allows our runtime to parse, execute, and return results seamlessly before the next reasoning step.  

% This explicit tagging helps the model generate syntactically correct SQL and reduces parsing errors. Figure~\ref{fig:tool_call_prompt} illustrates the full prompt template, including context framing, tool descriptions, and response formatting guidelines.

Then, how do LLMs actually invoke these tools? We adopt the standard multi-turn dialogue framework, where the model generates at least one tool call in each chat turn. Traditional tool-call interfaces generally restrict code parameters to a single quoted line, which can be cumbersome for complex SQL. In contrast, Qwen3 \cite{yang2025qwen3} extends this by allowing multi-line code snippets, demarcated with custom XML tags (\texttt{<code>…</code>}), to preserve formatting and readability \cite{NousResearch}. Drawing on this, we wrap each SQL query in a lightweight \texttt{<code></code>} block, providing the model with a clear, dedicated region for writing arbitrarily long, properly formatted statements. The following presents the full prompt schema, illustrating how reasoning turns and tagged tool-calls interleave.

\begin{AutoVerb}
\textcolor{red}{[System Prompt]}
You are Qwen, created by Alibaba Cloud. You are a helpful assistant.

# Tools

You may call one or more functions to assist with the user query.

You are provided with function signatures within <tools></tools> XML tags:
<tools>
\{"type": "function", "function": \{"name": "code_interpreter", "description": "SQL code interpreter", "parameters": null\}\}
\{"type": "function", "function": \{"name": "search", "description": "use key words to get a list of tables.", "parameters": \{"type": "object", "properties": \{"keywords": \{"type": "str", "description": "key words used for searching (sentence is accepted)."\}, "top_k": \{"type": "int", "description": "top k tables to return.", "default": 8\}\}, "required": ["keywords"]\}\}\}
</tools>

For each function call, return a json object with function name and arguments within <tool_call></tool_call> XML tags:
<tool_call>
\{"name": <function-name>, "arguments": <args-json-object>\}
</tool_call>
For code parameters, use placeholders first, and then put the code within <code></code> XML tags, such as:
<tool_call>
\{"name": <function-name>, "arguments": \{"code": ""\}\}
<code>
Here is the code.
</code>
</tool_call>

\textcolor{red}{[User Prompt]}
You are tasked with solving an open-domain table question answering problem. You must use tools to search for tables and execute SQL queries to find the answer. Follow the instructions below carefully:

- Always use the table name you searched as the target of your SQL queries.
- Set a proper `top_k` value to cover enough candidates and avoid omitting possible tables.
- Some tables have the similar schema with the same schema types. You can set proper `top_k` to access them and union them together help you better get the answer.
- Always perform both search and SQL steps, even if you "know" the answer, to ensure correctness.
- You may invoke tools in any order or combination until you have confidence in your result.
- The answer should be either a single item or a list of items. When you have the final answer, format it exactly as `<answer>item1,item2,…</answer>` with no extra words or punctuation.

**Question**: \textcolor{blue}{\{Question\}}
\end{AutoVerb}

\subsection{Adapted GRPO Algorithm}

Building on the impressive performance of Deepseek-R1, which popularized the use of the GRPO algorithm in open-domain reasoning tasks \cite{guo2025deepseek,shao2024deepseekmath}, we introduce several modifications to improve training efficiency and mitigate straggler effects in multi‐turn generation.

\paragraph{GRPO with Importance Sampling and Clipping}  
The GRPO loss incorporates importance sampling, a baseline, and a clipping mechanism to stabilize off‐policy updates under the Policy Gradient Theorem. For a rollout of $G$ episodes, it can be written as
\begin{gather}
    \mathcal{L}_{\mathrm{GRPO}}(\theta)
    = -\sum_{i=1}^G \sum_{t=1}^{T_i}
    \min\!\Bigl(r_{i,t}\,\hat{A}_i,\,
    \mathrm{clip}(r_{i,t},\,1-\varepsilon,\,1+\varepsilon)\,\hat{A}_i\Bigr), \\[4pt]
    r_{i,t} \;=\;\frac{\pi_\theta(o_{i,t}\mid q_i, o_{i,<t})}{\pi_{\mathrm{old}}(o_{i,t}\mid q_i, o_{i,<t})}, 
    \quad
    \hat{A}_i \;=\;\frac{R_i - \overline{R}}{\mathrm{std}(R_1,\dots,R_G)},
\end{gather}
where $R_i$ is the cumulative reward of episode $i$, and $\overline{R}$ its batch mean.  Thanks to importance weights $r_{i,t}$, samples generated by the old policy $\pi_{\mathrm{old}}$ remain useful for updating the current policy $\pi_\theta$, enabling an efficient off‐policy approach.

\paragraph{Rollout vs.\ Training Batches}  
In practice, we decouple data generation (rollout) from weight updates (training).  A ``rollout model" produces a large batch of samples—called the rollout batch—using an optimized inference engine (e.g.\ vLLM or SGLang), while a separate ``training model" consumes smaller training batches for gradient updates.  Once the training step completes with a rollout batch, its weights are synchronized back to the rollout model.  This separation accelerates sample generation but can introduce synchronization delays (straggler effects) when some rollout engines finish early and idle while waiting for the last, longer multi‐turn generations to complete.

\paragraph{Async GRPO with LoRA and Rollout Buffer}  
To alleviate idle time and maintain high GPU utilization, we propose \emph{Async GRPO}, which combines (1) lightweight LoRA adapters \cite{hu2022lora} and (2) a small \emph{rollout buffer}.  Since LoRA adapters contain only a fraction of the full model's parameters, many engines can load multiple adapters simultaneously.  Concretely:

\begin{enumerate}
  \item After each rollout batch, the training model updates its LoRA adapter parameters.
  \item New generations in the next rollout batch predominantly use the updated adapter.
  \item Meanwhile, the rollout buffer—a secondary, smaller batch—continues to generate samples with the \emph{previous} adapter.
  \item These buffered samples are then incorporated into the next training step, smoothing out variability in generation time.
\end{enumerate}

\begin{figure}[htbp]
    \centering
    \includegraphics[width=\textwidth]{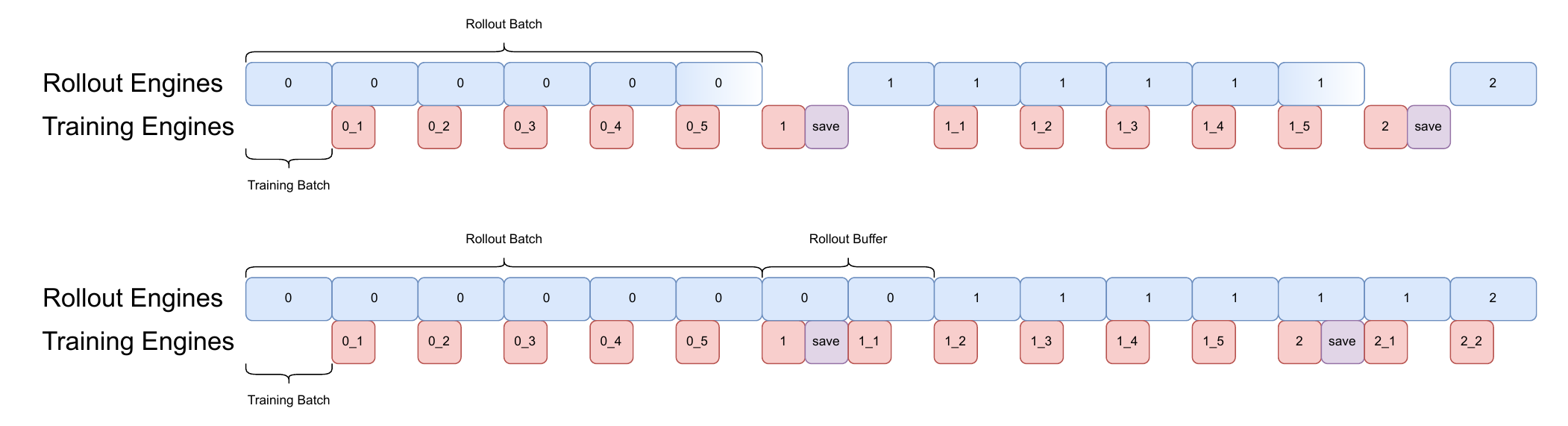}
    \caption{The time series of plain GRPO vs async GRPO. We illustrate the straggler effect at the end of every rollout batch with a gradient color.}
    \label{fig:comp_grpo}
\end{figure}

Figure~\ref{fig:comp_grpo} contrasts plain and Async GRPO time series.  We assume a 6:1 ratio between rollout and training batch sizes, and introduce a 2:1 buffer.  In plain GRPO, the final training batch often suffers severe straggler delays due to high variance in multi‐turn rollout times.  Async GRPO, by overlapping buffer generation with adapter updates, keeps all engines busy and effectively masks these delays.

% \section{Experiment}

% We select Open Wikitable as the main experiment dataset \cite{kweon2023open}. It contains a huge number of tables with many questions, which offer great advantages for us both in training and evaluation. The details of the Open WikiTable shows in Table \ref{tab:owtq_info}.

% \begin{table}[!htbp]
%   \centering
%   \begin{tabular}{ccccc}
%     \toprule
%     Part & Train & Valid & Test & Total\\
%     \midrule
%     Question & 53819 & 6602 & 6602 & 67023 \\
%     Table & 17275 & 2139 & 2262 & 21676 \\
%     \bottomrule
%   \end{tabular}
%   \caption{Data Information of the Open Wikitable Dataset.}
%   \label{tab:owtq_info}
% \end{table}

% Since the test set is way larger for the actual test for multi-turn generation. We randomly select 1024 samples from the 6602 test set. The selected data will replace the test set for all experiments.

% We organize our experiments as follows

% \begin{itemize}
%     \item \textbf{LLM Performance with Tool Call.} We lead a broad test with Qwen3 series to test model performance of open-domain table QA with the given tools.
%     \item \textbf{GRPO Training.} We use adapted GRPO algorithm to train a small model and test its performance.
%     % \item \textbf{Match Algorithm Influence.} We also tested the model performance with different \texttt{search} tool implementations.
% \end{itemize}

\section{Experiment}

\paragraph{Dataset}  
We conduct all experiments on the Open WikiTable dataset \cite{kweon2023open}, which comprises 67,023 question–table pairs drawn from English Wikipedia. Table~\ref{tab:owtq_info} summarizes the train/validation/test splits.

\begin{table}[!htbp]
  \centering
  \begin{tabular}{ccccc}
    \toprule
    \textbf{Split}    & \textbf{\#Questions} & \textbf{\#Tables} &  &  \\
    \midrule
    Train    & 53,819      & 17,275   &  &  \\
    Valid    &  6,602      &  2,139   &  &  \\
    Test     &  6,602      &  2,262   &  &  \\
    \midrule
    Total    & 67,023      & 21,676   &  &  \\
    \bottomrule
  \end{tabular}
  \caption{Overview of the Open WikiTable dataset.}
  \label{tab:owtq_info}
\end{table}

Because the full test split is relatively large for multi-turn tool‐call evaluation, we randomly sample 1,024 examples from the 6,602 test questions to form a held-out \emph{evaluation set}.  All models and configurations are evaluated on this same subset to ensure fair comparison.

\paragraph{Evaluation Metrics}  
We report three metrics on our 1,024-sample evaluation set:  
(1) \textbf{Exact Match (EM)}: the proportion of model outputs that exactly match the reference answer;  
(2) \textbf{Token Consumption}: the total number of tokens used, including prompt tokens, LLM responses, and any tool call outputs;  
(3) \textbf{Turn Count}: the number of dialogue turns required for the model to arrive at its final answer.

\paragraph{Experimental Setup}  
Our experiments consist of two parts:
\begin{itemize}
  \item \textbf{LLM Performance with Tool Call.}  
    We evaluate multiple Qwen3 variants using the same prompt templates and tool configurations described in Section~2.1.  Each model is tested on the held-out 1,024-sample subset, and we record EM, Token Consumption, and Turn Count to compare their zero-shot open-domain table QA capabilities.

  \item \textbf{Async GRPO Fine-Tuning.}  
    We begin with the Qwen3-4B base model and fine-tune it using our Async GRPO algorithm (Section~2.2).  After training, we evaluate the adapted model on the same 1,024-sample set and report its EM, Token Consumption, and Turn Count.
\end{itemize}

% \subsection{LLM Performance with Tool Call}

% In the first experiment, we compare different sizes of Qwen3 models with the default tool call implementations in the test set. We set the token limit to 16K with unlimited chat turns. Besides the accuracy of every model, we also count the average token consumption and chat turns. The result is in Table \ref{tab:llm_performance}.

% \begin{table}[!htbp]
%   \centering
%   \begin{tabular}{cccc}
%     \toprule
%     Model & Accuracy & Avg Token & Avg Turn \\
%     \midrule
%     Qwen3-4B & 12.01\% & 11997 & 52.99 \\
%     Qwen3-8B & 41.02\% & 5205 & 16.56 \\
%     Qwen3-32B & 58.20\% & 3918 & 8.98 \\
%     \bottomrule
%   \end{tabular}
%   \caption{Model Performances.}
%   \label{tab:llm_performance}
% \end{table}

% We can find that even with basic tool call implementations, the model over 8B can get a fairly good performance in the test set. However, when the model is smaller, it will spend more tokens and make more turns to figure out the answer. The very small model, Qwen3-4B, even have very large turn and token consumptions. We manually checked the model outputs and find that smaller model will have higher risk of repeating the chat content until it accesses the token limit.

% This result gives us a much more different insight of open-domain table QA. The mainstream solution mainly focuses on improving the match algorithm accuracy to make the search step get the correct table. However, with a sparse match algorithm and LLMs to manually adjust their keywords of match, they can still get good performance.

\subsection{LLM Performance with Tool Call}

We evaluate three Qwen3 variants on our 1,024-sample test split, using a 16K token window and unlimited dialogue turns. Table~\ref{tab:llm_performance} reports each model’s exact-match accuracy, average token consumption, and average turn count.

\begin{table}[!htbp]
  \centering
  \begin{tabular}{lccc}
    \toprule
    \textbf{Model}   & \textbf{Accuracy} & \textbf{Avg.\ Token} & \textbf{Avg.\ Turn} \\
    \midrule
    Qwen3-4B         & 12.0\%            & 11,997               &    52.9             \\
    Qwen3-8B         & 41.0\%            &  5,205               &    16.6             \\
    Qwen3-32B        & 58.2\%            &  3,918               &     9.0             \\
    \bottomrule
  \end{tabular}
  \caption{Zero-shot table QA performance of Qwen3 models with default tool calls.}
  \label{tab:llm_performance}
\end{table}

Several patterns emerge from these results. First, model size strongly correlates with accuracy: Qwen3-32B achieves nearly 60\% exact match, whereas Qwen3-4B attains only 12\%. Second, larger models require fewer tokens and dialogue turns, indicating more efficient reasoning and reduced reliance on repeated tool invocations. In contrast, the 4B-parameter model consumes substantially more tokens and turns—often repeating previous steps until it exhausts the context window. Manual inspection reveals that smaller models struggle to refine search keywords and tend to loop through similar queries.

These findings suggest that, beyond optimizing table retrieval algorithms, the inherent reasoning and planning capabilities of larger LLMs play a critical role in open-domain table QA. Even with a simple BM25+ search backend, high-capacity models can iteratively adjust their queries and achieve strong performance, while smaller models incur heavy overhead in token usage and dialogue complexity.

\subsection{GRPO Training}

In light of the zero-shot performance gap on Qwen3-4B, we employ a two-stage fine-tuning strategy: (1) a supervised \emph{cold start} on ``simple" questions, and (2) reinforcement learning with our Async GRPO algorithm on the remaining ``difficult" cases.

\paragraph{Data Partitioning and Cold Start}  
We first use the high-capacity Qwen3-32B model to generate answers for all 53,819 training questions.  Samples whose answers exactly match the ground truth are labeled \emph{simple} (31,959 items), while the rest are labeled \emph{difficult} (21,860 items), as shown in Table~\ref{tab:sd_dis}.  
\begin{table}[!htbp]
  \centering
  \begin{tabular}{lrrr}
    \toprule
    & \textbf{Simple} & \textbf{Difficult} & \textbf{Total} \\
    \midrule
    Count & 31,959 & 21,860 & 53,819 \\
    \bottomrule
  \end{tabular}
  \caption{Partition of training questions into simple and difficult subsets.}
  \label{tab:sd_dis}
\end{table}

We perform full parameter supervised fine-tuning (SFT) on the simple subset to provide a strong initialization for Qwen3-4B.  \emph{Cold start} hyperparameters include:
\begin{itemize}
  \item Learning rate: \(\eta=1\times10^{-5}\)
  \item Batch size: 32
  \item Epochs: 2
  \item Context window: 8K tokens
\end{itemize}
After the cold start, the model (denoted Qwen3-4B-ColdStart) achieves 72.1\% EM with an average of 3,906 tokens and 8.8 turns per example.

\paragraph{Async GRPO Fine-Tuning}  
Next, we apply our Async GRPO algorithm to the remaining difficult questions.  We attach a LoRA adapter of rank 12 to Qwen3-4B-ColdStart and train with the following hyperparameters:
\begin{itemize}
  \item Group value: $G=8$
  \item Rollout batch size: 512
  \item Training batch size: 64
  \item Rollout buffer size: 512
  \item Clipping parameter: \(\varepsilon = 0.2\)  
  \item AdamW optimizer, learning rate \(5\times10^{-5}\)  
  \item Total update steps: 1,504
  \item Context window: 8K tokens
\end{itemize}
After Async GRPO, the model surpasses its cold-start performance, achieving 86.2\% EM with only 2,690 tokens and 6.8 turns on average.  For reference, Table~\ref{tab:grpo_result} also reports the zero-shot and cold-start baselines as well as the upper–bound Qwen3-8B and Qwen3-32B models.

\begin{table}[!htbp]
  \centering
  \begin{tabular}{lccc}
    \toprule
    \textbf{Model}               & \textbf{Accuracy} & \textbf{Avg.\ Token} & \textbf{Avg.\ Turn} \\
    \midrule
    Qwen3-4B (Zero-Shot)         & 12.0\%            & 11,997               & 52.9                \\
    Qwen3-4B (Cold Start)     & 72.1\%            &  3,906               &  8.8                \\
    Qwen3-4B (Cold Start+Async GRPO) & 86.2\%            &  2,690               &  6.8                \\
    \midrule
    Qwen3-8B (Zero-Shot)         & 41.0\%            &  5,205               & 16.6                \\
    Qwen3-32B (Zero-Shot)        & 58.2\%            &  3,918               &  9.0                \\
    \bottomrule
  \end{tabular}
  \caption{Performance comparison after each training stage.}
  \label{tab:grpo_result}
\end{table}

These results confirm that (i) a supervised cold-start on simple examples establishes a robust initial policy—yielding a more than 60\% jump in exact-match accuracy and dramatic reductions in token and turn counts—and (ii) Async GRPO further hones the model on difficult cases, delivering an additional 14\% absolute gain in accuracy while cutting token consumption and dialogue complexity by roughly 30\%.

Altogether, our two-stage pipeline—cold-start SFT followed by off-policy Async GRPO—demonstrates that reinforcement learning can act as a critical corrective mechanism under tool-call constraints. By combining stability from supervised fine-tuning with the exploration efficiency of importance-sampled policy updates, we obtain a compact 4B-parameter agent whose end-to-end performance rivals that of much larger LLMs, making scalable open-domain table QA both accurate and computationally efficient.

% \subsection{Match Algorithm Influence}

% In this experiment, we will change the backend implementation of the \texttt{search} tool. We will use the OpenTable-R1  trained in the previous experiment.

% \section{Related Work}

\section{Related Work}

\paragraph{Open-Domain Table Question Answering}  
Early work in table QA focused on semantic parsing over a single table, with systems such as TaPas \cite{herzig2020tapas} and TAPEX \cite{liu2021tapex} demonstrating strong closed-domain performance. Open-domain table QA—in which a large corpus of tables must be searched before answering—raises additional retrieval challenges. Kweon \emph{et al.} introduced the Open WikiTable dataset to benchmark end-to-end retrieval and QA over Wikipedia tables \cite{kweon2023open}. Traditional pipelines rely on static retrieval (e.g.\ TF–IDF, BM25) followed by closed-domain parsing \cite{liu2021tapex}, but recent work has begun to integrate retrieval and reasoning in a single loop.

\paragraph{Tool-Augmented LLMs}  
The ability of large language models to invoke external tools—search engines, calculators, or code interpreters—has become a popular mechanism to extend their capabilities \cite{schick2023toolformer}. Toolformer automatically learns when to call APIs during pretraining \cite{schick2023toolformer}, while Qwen3 employs a Hermes-style XML tagging scheme to integrate multi-turn tool calls for tasks including table QA \cite{yang2025qwen3,NousResearch}. Our work builds on this agentic framework, exposing simple \texttt{search} and \texttt{code\_interpreter} APIs to steer LLM reasoning over large table corpora.

\paragraph{Retrieval Methods}  
Classical sparse retrieval (BM25, TF–IDF) remains a strong baseline for table indexing \cite{robertson2009probabilistic}, but dense passage retrieval (DPR) and neural ranking models often yield higher recall at scale \cite{karpukhin2020dense}. Recent studies combine sparse and dense signals in hybrid indexes \cite{zheng2022sparta}, and continuous retrieval via end-to-end retriever–reader training. We adopt BM25+ for simplicity, but our modular design allows future integration of neural table retrievers.

\paragraph{Reinforcement Learning and Policy Fine-Tuning}  
Reinforcement learning from human feedback (RLHF) and policy gradient methods have been used to refine LLM behavior \cite{ouyang2022training}. In particular, off-policy algorithms with importance sampling and clipping—such as PPO \cite{schulman2017proximal} and our GRPO variant—stabilize updates while leveraging previously collected trajectories. Adapter-based methods like LoRA \cite{hu2022lora} and AdaLoRA \cite{zhang2023adalora} reduce tuning costs by injecting low-rank updates. Our Async GRPO further addresses straggler effects in multi-turn rollout by overlapping buffer generation with adapter updates.

\paragraph{Asynchronous and Efficient RL}  
Several works have explored asynchronous actor-critic architectures to improve throughput, notably A3C/A2C \cite{mnih2016asynchronous} in RL for games and robotics. In LLM fine-tuning, recent pipelines decouple sample generation from updates (e.g.\ DeepSeek’s retrieval-guided RL \cite{guo2025deepseek}), but few address multi-turn variance in rollout time. Our rollout-buffer mechanism draws inspiration from asynchronous training frameworks to maximize hardware utilization under tool-call constraints.

\paragraph{Summary}  
Our approach uniquely combines multi-turn tool calls, supervised cold starts, and asynchronous off-policy RL to train a compact table QA agent. By bridging retrieval, reasoning, and efficient fine-tuning, we advance the state of the art in building scalable, accurate, and resource-efficient open-domain table QA systems.

% \section{Conclusion}

% In this paper, we introduced an end-to-end tool call based open-domain table QA agent. By utilizing the current model's tool call ability, we only need to implement specific tools for the open-domain table QA task. The LLMs can perform well with these tools. Based on this feature, we introduce GRPO training to reinforce the model that does not perform well. The model can get very large boost.

\section{Conclusion}

In this work, we presented a practical framework for open-domain table question answering that leverages multi-turn tool calls to external search and SQL execution engines.  We demonstrated that even with a simple BM25+ retrieval backend and basic tool APIs, large-capacity LLMs (e.g.\ Qwen3-32B) can achieve strong zero-shot performance by iteratively refining their queries, while smaller models struggle under the same setup.

To close this gap, we proposed a two-stage fine-tuning pipeline for a compact 4B-parameter model.  First, a supervised cold-start on ``easy" examples provided a reliable initialization, boosting exact-match accuracy from 12\% to over 70\% and sharply reducing token and turn usage.  Second, our Async GRPO algorithm—with importance-sampled off-policy updates, LoRA adapters, and a rollout buffer—further refined the model on ``difficult" cases.  The resulting agent achieved 86\% EM while consuming less than 3K tokens and under 7 turns per query, rivaling larger zero-shot baselines.

Our findings underscore two key insights: (1) retrieval quality alone does not determine open-domain table QA success; model reasoning and planning capabilities are equally crucial, and (2) reinforcement learning—with careful batching and asynchronous rollout strategies—can dramatically improve throughput and final accuracy of compact agents in tool-call settings.

Looking ahead, we plan to explore more advanced retrieval methods and richer tool interfaces. We believe that combining structured tool calls with targeted RL fine-tuning will continue to be a promising direction for building accurate, efficient, and scalable table QA systems.

\bibliographystyle{unsrt}
\bibliography{ref}

\begin{thebibliography}{10}

\bibitem{grattafiori2024llama}
Aaron Grattafiori, Abhimanyu Dubey, Abhinav Jauhri, Abhinav Pandey, Abhishek Kadian, Ahmad Al-Dahle, Aiesha Letman, Akhil Mathur, Alan Schelten, Alex Vaughan, et~al.
\newblock The llama 3 herd of models.
\newblock {\em arXiv preprint arXiv:2407.21783}, 2024.

\bibitem{qwen2025qwen25technicalreport}
Qwen, :, An~Yang, Baosong Yang, Beichen Zhang, Binyuan Hui, Bo~Zheng, Bowen Yu, Chengyuan Li, Dayiheng Liu, Fei Huang, Haoran Wei, Huan Lin, Jian Yang, Jianhong Tu, Jianwei Zhang, Jianxin Yang, Jiaxi Yang, Jingren Zhou, Junyang Lin, Kai Dang, Keming Lu, Keqin Bao, Kexin Yang, Le~Yu, Mei Li, Mingfeng Xue, Pei Zhang, Qin Zhu, Rui Men, Runji Lin, Tianhao Li, Tianyi Tang, Tingyu Xia, Xingzhang Ren, Xuancheng Ren, Yang Fan, Yang Su, Yichang Zhang, Yu~Wan, Yuqiong Liu, Zeyu Cui, Zhenru Zhang, and Zihan Qiu.
\newblock Qwen2.5 technical report, 2025.

\bibitem{yang2025qwen3}
An~Yang, Anfeng Li, Baosong Yang, Beichen Zhang, Binyuan Hui, Bo~Zheng, Bowen Yu, Chang Gao, Chengen Huang, Chenxu Lv, et~al.
\newblock Qwen3 technical report.
\newblock {\em arXiv preprint arXiv:2505.09388}, 2025.

\bibitem{schulman2017proximal}
John Schulman, Filip Wolski, Prafulla Dhariwal, Alec Radford, and Oleg Klimov.
\newblock Proximal policy optimization algorithms.
\newblock {\em arXiv preprint arXiv:1707.06347}, 2017.

\bibitem{rafailov2023direct}
Rafael Rafailov, Archit Sharma, Eric Mitchell, Christopher~D Manning, Stefano Ermon, and Chelsea Finn.
\newblock Direct preference optimization: Your language model is secretly a reward model.
\newblock {\em Advances in Neural Information Processing Systems}, 36:53728--53741, 2023.

\bibitem{shao2024deepseekmath}
Zhihong Shao, Peiyi Wang, Qihao Zhu, Runxin Xu, Junxiao Song, Xiao Bi, Haowei Zhang, Mingchuan Zhang, YK~Li, Y~Wu, et~al.
\newblock Deepseekmath: Pushing the limits of mathematical reasoning in open language models.
\newblock {\em arXiv preprint arXiv:2402.03300}, 2024.

\bibitem{NousResearch}
NousResearch.
\newblock Nousresearch/hermes-function-calling.

\bibitem{guo2025deepseek}
Daya Guo, Dejian Yang, Haowei Zhang, Junxiao Song, Ruoyu Zhang, Runxin Xu, Qihao Zhu, Shirong Ma, Peiyi Wang, Xiao Bi, et~al.
\newblock Deepseek-r1: Incentivizing reasoning capability in llms via reinforcement learning.
\newblock {\em arXiv preprint arXiv:2501.12948}, 2025.

\bibitem{hu2022lora}
Edward~J Hu, Yelong Shen, Phillip Wallis, Zeyuan Allen-Zhu, Yuanzhi Li, Shean Wang, Lu~Wang, Weizhu Chen, et~al.
\newblock Lora: Low-rank adaptation of large language models.
\newblock {\em ICLR}, 1(2):3, 2022.

\bibitem{kweon2023open}
Sunjun Kweon, Yeonsu Kwon, Seonhee Cho, Yohan Jo, and Edward Choi.
\newblock Open-wikitable: Dataset for open domain question answering with complex reasoning over table.
\newblock {\em arXiv preprint arXiv:2305.07288}, 2023.

\bibitem{herzig2020tapas}
Jonathan Herzig, Pawe{\l}~Krzysztof Nowak, Thomas M{\"u}ller, Francesco Piccinno, and Julian~Martin Eisenschlos.
\newblock Tapas: Weakly supervised table parsing via pre-training.
\newblock {\em arXiv preprint arXiv:2004.02349}, 2020.

\bibitem{liu2021tapex}
Qian Liu, Bei Chen, Jiaqi Guo, Morteza Ziyadi, Zeqi Lin, Weizhu Chen, and Jian-Guang Lou.
\newblock Tapex: Table pre-training via learning a neural sql executor.
\newblock {\em arXiv preprint arXiv:2107.07653}, 2021.

\bibitem{schick2023toolformer}
Timo Schick, Jane Dwivedi-Yu, Roberto Dess{\`\i}, Roberta Raileanu, Maria Lomeli, Eric Hambro, Luke Zettlemoyer, Nicola Cancedda, and Thomas Scialom.
\newblock Toolformer: Language models can teach themselves to use tools.
\newblock {\em Advances in Neural Information Processing Systems}, 36:68539--68551, 2023.

\bibitem{robertson2009probabilistic}
Stephen Robertson, Hugo Zaragoza, et~al.
\newblock The probabilistic relevance framework: Bm25 and beyond.
\newblock {\em Foundations and Trends{\textregistered} in Information Retrieval}, 3(4):333--389, 2009.

\bibitem{karpukhin2020dense}
Vladimir Karpukhin, Barlas Oguz, Sewon Min, Patrick~SH Lewis, Ledell Wu, Sergey Edunov, Danqi Chen, and Wen-tau Yih.
\newblock Dense passage retrieval for open-domain question answering.
\newblock In {\em EMNLP (1)}, pages 6769--6781, 2020.

\bibitem{zheng2022sparta}
Ningxin Zheng, Bin Lin, Quanlu Zhang, Lingxiao Ma, Yuqing Yang, Fan Yang, Yang Wang, Mao Yang, and Lidong Zhou.
\newblock $\{$SparTA$\}$:$\{$Deep-Learning$\}$ model sparsity via $\{$Tensor-with-Sparsity-Attribute$\}$.
\newblock In {\em 16th USENIX Symposium on Operating Systems Design and Implementation (OSDI 22)}, pages 213--232, 2022.

\bibitem{ouyang2022training}
Long Ouyang, Jeffrey Wu, Xu~Jiang, Diogo Almeida, Carroll Wainwright, Pamela Mishkin, Chong Zhang, Sandhini Agarwal, Katarina Slama, Alex Ray, et~al.
\newblock Training language models to follow instructions with human feedback.
\newblock {\em Advances in neural information processing systems}, 35:27730--27744, 2022.

\bibitem{zhang2023adalora}
Qingru Zhang, Minshuo Chen, Alexander Bukharin, Nikos Karampatziakis, Pengcheng He, Yu~Cheng, Weizhu Chen, and Tuo Zhao.
\newblock Adalora: Adaptive budget allocation for parameter-efficient fine-tuning.
\newblock {\em arXiv preprint arXiv:2303.10512}, 2023.

\bibitem{mnih2016asynchronous}
Volodymyr Mnih, Adria~Puigdomenech Badia, Mehdi Mirza, Alex Graves, Timothy Lillicrap, Tim Harley, David Silver, and Koray Kavukcuoglu.
\newblock Asynchronous methods for deep reinforcement learning.
\newblock In {\em International conference on machine learning}, pages 1928--1937. PmLR, 2016.

\end{thebibliography}

\clearpage
%\appendix

%\section{Appendix 1}

%%%%%%%%%%%%%%%%%%%%%%%%%%%%%%%%%%%%%%%%%%%%%%%%%%%%%%%%%%%%

%\input{checklist}

\end{document}